\documentstyle[ijcai95]{article}

\newcommand{\lc}{\rule[1.2ex]{.05em}{.24em}^{\hspace*{-0.05em}\rule[.6ex]{.25em}{.05em}}\!}
\newcommand{\rc}{\:\rule[1.2ex]{.05em}{.24em}^{\hspace*{-.26em}\rule[.6ex]{.25em}{.05em}}}

\input{psfig}

\begin{document}

\title{The Use of Knowledge Preconditions in Language
Processing\thanks{This work was done as part of my dissertation
research at Harvard University, and was supported by a Bellcore
graduate fellowship and by U S WEST Advanced Technologies.  I would
like to thank Barbara Grosz, Stuart Shieber, and Candy Sidner for
their helpful comments, discussions, and insights on this work.}}

\author{Karen E. Lochbaum\\
U S WEST Advanced Technologies\\
4001 Discovery Dr.\\
Boulder, CO  80303\\
klochba@advtech.uswest.com}

\maketitle

\begin{abstract}
If an agent does not possess the knowledge needed to perform an
action, it may privately plan to obtain the required information on
its own, or it may involve another agent in the planning process by
engaging it in a dialogue.  In this paper, we show how the
requirements of knowledge preconditions can be used to account for
information-seeking subdialogues in discourse.  We first present an
axiomatization of knowledge preconditions for the SharedPlan model of
collaborative activity \cite{Grosz-Kraus:93}, and then provide an
analysis of information-seeking subdialogues within a general
framework for discourse processing.  In this framework, SharedPlans
and relationships among them are used to model the intentional
component of Grosz and Sidner's~\shortcite{Grosz-Sidner:86} theory of
discourse structure.
\end{abstract}

\vspace*{-2.75ex}
\section{Introduction}
For an agent\footnote{Unless otherwise indicated, we will use the term
``agent'' to refer to both individual agents and sets of agents.}  to
be able to perform an action, it must satisfy both the physical and
knowledge preconditions of that action \cite{Moore:85,Morgenstern:87}.
For example, for an agent to pick up a particular tower of blocks, it
must (1)~know how to pick up towers in general, (2)~be able to
identify the tower in question, and (3)~ have satisfied the (physical)
preconditions or constraints associated with picking up towers (e.g.,
it must have a free hand).  These conditions must hold whether the
agent is planning an action on its own or is involved in a
collaborative planning effort with other agents.

In this paper, we provide an axiomatization of knowledge preconditions
for the SharedPlan model of collaborative activity
\cite{Grosz-Sidner:90,LGS:90,Grosz-Kraus:93}.  This model draws upon
past work \cite{Moore:85,Morgenstern:87}, but adapts it to the
collaborative situation.  We briefly describe the SharedPlan framework
in Section~\ref{sp}, and then, in Section~\ref{know-prec}, present our
axiomatization of knowledge preconditions.  In Section~\ref{role}, we
demonstrate the use of knowledge preconditions in accounting for
information-seeking subdialogues, such as those in Figure~\ref{ac}.  We
then compare our approach to the alternative accounts
\cite{Litman-Allen:87,Lambert-Carberry:91,Ramshaw:91}.

\begin{small}
\begin{figure}[hbtp]
\hspace*{0.2in}
\psfig{figure=knowp-d.eps,width=3.1in}
\caption{\small Information-Seeking Subdialogues [Grosz, 1974]\label{ac}}
\end{figure}
\end{small}

\vspace*{-3ex}
\section{SharedPlans}
\label{sp}
The SharedPlan formalism is a mental-state model of collaborative
plans with roots in Pollack's~\shortcite{Pollack:90} work on
single-agent plans.  For a group of agents $GR$ to have a {\em full
SharedPlan} (FSP) for an act $\alpha$, they must satisfy the
requirements given in Figure~\ref{sp-reqts}.  When the agents have
satisfied only a subset of these requirements, they are said to have a
{\it partial SharedPlan} (PSP).\footnote{This description of a PSP is
only a rough, though useful, approximation to the formal definition
given by Grosz and Kraus~\shortcite{Grosz-Kraus:93}.}  The bracketed
terms in Figure~\ref{sp-reqts} indicate the operators used by Grosz
and Kraus~\shortcite{Grosz-Kraus:93} to formalize each requirement.

Requirement~(1) in Figure~\ref{sp-reqts} refers to the agents' {\em
recipe} \cite{Pollack:90} for $\alpha$.  Recipes are modeled in Grosz
and Kraus's definitions as sets of constituent acts and constraints.
To perform an act $\alpha$, an agent must perform each constituent act
in $\alpha$'s recipe according to the constraints of that recipe.
Actions themselves may be further decomposed into act-types and
parameters.  We will represent an action $\alpha$ as a term of the
form $\bar{\alpha}(p_1,\ldots,p_n)$ where $\bar{\alpha}$ represents
the act-type of the action and the $p_i$ its parameters.


\begin{figure}[t]
\begin{small}
For a group of agents $GR$ to have an FSP for $\alpha$
\begin{enumerate}
\item $GR$ must have mutual belief of a {\em recipe} for $\alpha$
\item For each single-agent constituent act of the recipe, there must
be an agent $G_{\beta_i} \in GR$, such that
\begin{enumerate}
\vspace*{-1ex}
\item $G_{\beta_i}$ intends to perform $\beta_i$ {\bf [Int.To]}
\item $G_{\beta_i}$ believes that it can perform $\beta_i$ {\bf [BCBA]}
\item $G_{\beta_i}$ has a full individual plan for $\beta_i$ {\bf [FIP]}
\item The group $GR$ has mutual belief of (2a)-(2c)
\item Each member of $GR$ intends that $G_{\beta_i}$ succeed {\bf [Int.Th]}
\end{enumerate}

\item For each multi-agent constituent act of the recipe, there must
be a subgroup of agents $GR_{\beta_i} \subseteq GR$ such that
\begin{enumerate}
\item $GR_{\beta_i}$ mutually believe that
they can perform $\beta_i$ {\bf [MBCBAG]}
\item $GR_{\beta_i}$ has a full SharedPlan for $\beta_i$ {\bf [FSP]}
\item The group $GR$ has mutual belief of (3a)-(3b)
\item Each member of $GR$ intends that $GR_{\beta_i}$ succeed {\bf [Int.Th]}
\end{enumerate}
\end{enumerate}
\vspace*{-2ex}
\caption{\small FSP Requirements \label{sp-reqts}}
\end{small}
\end{figure}

\section{Knowledge Preconditions}
\label{know-prec}
Grosz and Kraus~\shortcite{Grosz-Kraus:93} use the operators BCBA
(read ``believes can bring about'') and MBCBAG (read ``mutually
believe can bring about group'') to formalize respectively
requirements~(2b) and~(3a) in Figure~\ref{sp-reqts}.  Although these
operators are intended to specify the conditions under which an agent
is able to perform an action, their definitions explicitly require
only that an agent satisfy the physical preconditions or constraints
associated with an action to be able to perform it.  Because an agent
is not truly capable of performing an act unless it possesses the
appropriate knowledge, the definitions of BCBA and MBCBAG must be
augmented with an axiomatization of knowledge preconditions.  The
following observations made by Morgenstern~\shortcite{Morgenstern:87},
but recast in our terminology, must be represented in such an
axiomatization:
\begin{enumerate}
\vspace*{-1ex}
\item Agents need to know recipes for the acts they
perform.
\vspace*{-1.5ex}
\item All agents have some primitive acts in their
repertoire.
\vspace*{-1.5ex}
\item Agents must be able to identify the parameters of the acts they
perform.
\vspace*{-1.5ex}
\item Agents may know only some descriptions of an act.
\vspace*{-3.5ex}
\item Agents know that the knowledge necessary for complex acts
derives from that necessary for their component acts.
\vspace*{-1.5ex}
\end{enumerate}

Our axiomatization of knowledge preconditions is based on
Morgenstern's observations, but adapted to the requirements of
individual and shared mental-state plans.\footnote{A comparison of our
formalization with those of Morgenstern~\shortcite{Morgenstern:87} and
Moore~\shortcite{Moore:85} can be found elsewhere \cite{KEL:94}.}  We
use the predicates {\em has.recipe\/} and {\em id.params\/} to
represent explicitly observations~(1) and (3) above.  The remaining
observations are implicitly represented by the way in which these two
knowledge precondition relations are defined.  Observation~(2) is
modeled as the base case of {\em has.recipe}, and observation~(5) is
modeled by the use of {\em has.recipe\/} within the recursive plan
definitions.

Observation~(4) requires that the knowledge precondition relations be
intensional, rather than extensional; within their scope it should not
be possible to freely substitute one representation of an action for
another.  We thus define {\em has.recipe\/} and {\em id.params\/} to
hold of action {\em descriptions}, rather than actions.  Action
descriptions are intensional objects; one action description can be
substituted for another only if the descriptions are the same.  For
example, although $555$-$1234$ and $phone$-$number(speech$-$lab)$ may
be extensionally equivalent, the descriptions $\lc\,{555}$-${1234}\rc$
and $\lc{phone}$-${number(speech}$-${lab)}\rc$ are not.  By
convention, we will omit the corner quote notation in what follows and
simply take the appropriate arguments of the predicates to represent
action descriptions rather than actions.


Although Morgenstern's observations are most naturally expressed
informally in terms of knowledge, we formalize them using belief to
allow for the possibility of an agent's being incorrect.  Although it
is true that an agent cannot successfully act unless its beliefs about
recipes and parameters are correct, having to {\it know} the recipes
and parameters is too strong a requirement for collaborating agents
\cite{KEL:94}.

\subsection{Determining Recipes: {\em has.recipe}}
\label{recipe}
For an agent to be able to perform an act $\alpha$, it must know {\em
how\/} to perform $\alpha$; i.e., it must have a recipe for the act.
The relation $has.recipe(G,\alpha,R,T)$ is used to represent that
agent $G$ has a recipe $R\,$ for an act $\alpha$ at time $T$.  It is
formalized as follows:
\begin{footnotesize}
\begin{tabbing}
{}~~~~~~~~$has.recipe(G,\alpha,R,T) \Leftrightarrow$\\
{}~~~~~~~~(1)~~~~$[$\=$basic.level(\alpha)~~\wedge$\\
       \>$BEL(G,basic.level(\alpha),T) \wedge R=R_{Empty}]~~\vee$ \\
{}~~~~~~~~(2)~~~~$[\neg basic.level(\alpha)~~\wedge$\\
{}~~~~~~~~(2a)   \>$R=\{\beta_i,\rho_j\}~~\wedge$\\
{}~~~~~~~~(2a1)   \>~~~~$\{$\=$[|G| = 1 \wedge BEL(G,R \in
Recipes(\alpha),T)]~~\vee$\\
{}~~~~~~~~(2a2)       \>\>$[|G| > 1 \wedge MB(G,R \in Recipes(\alpha),T)]\}]$
\end{tabbing}
\end{footnotesize}

Clause~(1) of the definition models Morgenstern's second observation,
namely that agents do not need a recipe to perform a {\em basic-level\/}
action, i.e., one executable at will
\cite{Pollack:90}.\footnote{Basic-level actions are by their
nature single-agent actions.} For non-basic-level actions
(Clause~(2)), the agent of $\alpha$ (either a single agent (2a1) or a
group of agents (2a2)) must believe that some set of acts, $\beta_i$,
and constraints, $\rho_j$, constitute a recipe for
$\alpha$.

\subsection{Identifying Parameters: {\em id.params}}
\label{params}
An agent must also be able to identify the parameters of an act
$\alpha$ to be able to perform it.  For example, if an agent is told
{\it ``remove the flywheel,''} as in the dialogue of Figure~\ref{ac},
the agent must be able to identify the flywheel in question.  The
relation $id.params(G,\alpha,T)$ is used to represent that agent $G$
can identify the parameters of act $\alpha$ at time $T$.  If $\alpha$
is of the form $\bar{\alpha}(p_1,...,p_n)$, then
$id.params(G,\alpha,T)$ is true if $G$ can identify each of the $p_i$.
To do so, $G$ must have a description of each $p_i$ that is suitable
for $\bar{\alpha}$.  The relation {\em id.params\/} is defined as
follows:
\begin{footnotesize}
\begin{tabbing}
{}~~~~$id.params(G,\bar{\alpha}(p_1,\ldots,p_n),T) \Leftrightarrow$\\
{}~~~~~~~$(\forall i, 1 \leq i \leq n)~has.sat.descr(G,p_i,{\cal
F}(\bar{\alpha},p_i),T)$
\end{tabbing}
\end{footnotesize}

The ability to identify an object is highly context dependent.  For
example, as Appelt points out~\shortcite[pg. 200]{Appelt:85}, ``the
description that one must know to carry out a plan requiring the
identification of `John's residence' may be quite different depending
on whether one is going to visit him, or mail him a letter.''  The
function ${\cal F}$ in the above definition is an oracle function
intended to model the context-dependent nature of parameter
identification.  This function returns a suitable {\em identification
constraint\/} \cite{Appelt-Kronfeld:87} for a parameter $p_i$ in the
context of an act-type $\bar{\alpha}$.  For example, in the case of
sending a letter to John's residence, the constraint produced by the
oracle function would be that John's residence be described by a
postal address.

The relation $has.sat.descr(G,P,C,T)$ holds of an agent $G$, a
parameter description $P$, an identification constraint $C$, and a
time $T$, if $G$ has a suitable description, as determined by $C$, of
the object described as $P$ at time $T$.  To formalize this relation,
we utilize Kronfeld's~\shortcite{Kronfeld:86} notion of an
individuating set.  An agent's individuating set for an object is a
maximal set of terms such that each term is believed by the agent to
denote that object.  For example, an agent's individuating set for
John's residence might include its postal address as well as an
identifying physical description such as ``the only yellow house on
Cherry Street.''  To model individuating sets, we introduce a function
$IS(G,P,T)$; the function returns an agent $G$'s individuating set at
time $T$ for the object that $G$ believes can be described as $P$.
This function is based on similar elements of the formal language that
Appelt and Kronfeld~\shortcite{Appelt-Kronfeld:87} introduce as part
of their theory of referring.  The function returns a set that
contains $P$ as well as the other descriptions that $G$ has for the
object that it believes $P$ denotes.

For an agent to suitably identify a parameter described as $P$, the
agent must have a description, $P^\prime$, of the parameter such that
$P^\prime$ is of the appropriate sort.  For example, for an agent to
visit John's residence, it is not sufficient for the agent to believe
that the description ``John's residence'' refers to the place where
John lives.  Rather, the agent needs another description of John's
residence, one such as ``the only yellow house on Cherry Street,''
that is appropriate for the purpose of visiting him.  To model an
agent's ability to identify a parameter (described as $P$) for some
purpose, we thus require that the agent have an individuating set for
the parameter that contains a description $P^\prime$ such that
$P^\prime$ satisfies the identification constraint that derives from
the purpose.  The definition of {\em has.sat.descr\/} is thus as
follows:\footnote{A more precise account of what it means to be able
to identify an object is beyond the scope of this paper; for further
details, see the discussions by Hobbs~\shortcite{Hobbs:85}, Appelt and
Kronfeld~\shortcite{Appelt:85,Kronfeld:86,Appelt-Kronfeld:87}, and
Morgenstern~\shortcite{Morgenstern-thesis}.}
\begin{footnotesize}
\begin{tabbing}
{}~~~~$has.sat.descr(G,P,C,T) \Leftrightarrow$\\
{}~~~~~~$\{$\=$[$\=$|G| = 1~\wedge$\\
          \>\>~~$(\exists P^\prime) BEL(G,[$\=$P^\prime \in
IS(G,P,T)~~\wedge$\\
          \>\>\>${\it suff\!.for.id}(C,P^\prime)],T)]~~\vee$\\[0.5ex]
        \>$[|G| > 1~\wedge$\\
        \>\>~~$(\exists P^\prime) MB(G,(\forall G_j \in G)
[$\=$P^\prime \in IS(G_j,P,T)~~\wedge$\\
                           \>\>\>${\it suff\!.for.id}(C,P^\prime)],T)]\}$
\end{tabbing}
\end{footnotesize}
The predicate ${\it suff\!.for.id}(C,P^\prime)$ is true if the
constraint $C$ applies to the parameter description $P^\prime$.  The
oracle function ${\cal F}(\bar{\alpha},p_i)$ in {\em id.params\/}
produces the appropriate identification constraint on $p_i$ given
$\bar{\alpha}$.


\section{The Role of Knowledge Preconditions in Language Processing}
\label{role}
We now show how the requirements of knowledge preconditions can be
used in discourse processing.  Our model of discourse processing is
based on the theory of discourse structure proposed by Grosz and
Sidner~\shortcite{Grosz-Sidner:86}.  According to their theory,
discourse structure consists of three interrelated components: a
linguistic structure, an attentional state, and an intentional
structure.  The linguistic structure consists of discourse
segments\footnote{The term discourse segment is a generalization of
the term subdialogue.  Whereas the term discourse segment applies to
all types of discourse, the term subdialogue is reserved for segments
that occur within dialogues.} and an embedding relationship among
them; the bold rule in Figure~\ref{ac} indicates the linguistic
structure of that discourse.  Attentional state is an abstraction of
the discourse participants' focus of attention; it serves as a record
of those entities that are salient at any point in the discourse.
Intentional structure is comprised of discourse segment purposes and
their interrelationships, particularly that of {\em dominance}.  A
discourse segment purpose, or DSP, is a
Gricean-like~\shortcite{Grice:69} intention that leads to the
initiation of a discourse segment.  One DSP is dominated by another if
the satisfaction of the first provides part of the satisfaction of the
second.

Intentional structure plays a central role in discourse processing; an
agent's comprehension of the utterances in a discourse relies on the
recognition of this structure \cite{Grosz-Sidner:86}.  Grosz and
Sidner~\shortcite{Grosz-Sidner:90} proposed SharedPlans to provide a
basis for recognizing intentional structure.  They argued that
discourses are fundamentally collaborative, and hence that a model of
{\it shared\/} plans provides a more appropriate basis for discourse
processing than a model of single-agent plans.  However, the
connection between SharedPlans and intentional structure was never
specified.

\subsection{SharedPlans as Intentional Structure}
We have developed a model of discourse processing that provides that
connection \cite{KEL:94}.  Figure~3 illustrates the role of
SharedPlans in modeling intentional structure.  Each segment of a
discourse has an associated SharedPlan.  The purpose of the segment is
taken to be intention that (Int.Th \cite{Grosz-Kraus:93}) the
discourse participants form that plan.  This intention is held by the
agent who initiates the segment.  In what follows, we will refer to
that participant as the initiating conversational participant or ICP;
the other participant is the OCP \cite{Grosz-Sidner:86}.  Dominance
relationships between DSPs are modeled using {\it subsidiary}
relationships between SharedPlans.  One plan is subsidiary to another
if the completion of the first plan contributes to the completion of
the second.  Subsidiary relationships are discussed in more detail in
Section~\ref{recog}

The utterances of a discourse are understood in terms of their
contribution to the SharedPlans associated with the segments of the
discourse.  Those segments that have been completed at the time of
processing an utterance have a full SharedPlan (FSP) associated with
them (e.g., segment~(2) in Figure~3), while those that have not have a
partial SharedPlan (PSP) (e.g., segments~(1) and~(3) in Figure~3).

\psfig{figure=struct-ijcai.eps,width=3.37in}
\begin{center}
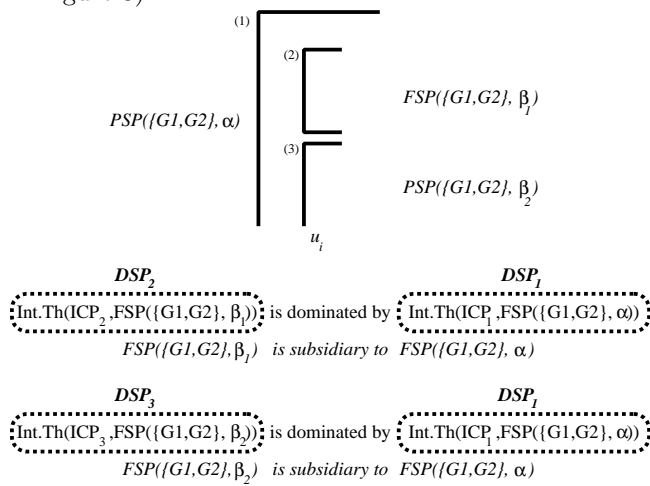
Figure 3: Modeling Intentional Structure
\end{center}

For each utterance of a discourse, an agent must determine whether the
utterance begins a new segment of the discourse, contributes to the
current segment, or completes it \cite{Grosz-Sidner:86}.  For an
utterance to begin a new segment, it must indicate the initiation of a
subsidiary plan.  This case is described in further detail below.  For
an utterance to contribute to the current segment, it must advance the
partial SharedPlan associated with the segment towards completion.
That is, it must establish one of the beliefs or intentions required
for the discourse participants to have a full SharedPlan, but missing
from their current partial SharedPlan.  For an utterance to complete
the current segment, it must indicate that the purpose of that segment
has been satisfied.  For that to be the case, the SharedPlan
associated with the segment must be an FSP rather than a PSP.  That
is, all of the beliefs and intentions required of an FSP, as indicated
in Figure~\ref{sp-reqts}, must have been established over the course
of the segment.

A detailed description of the implemented algorithms used in modeling
each of these cases can be found elsewhere \cite{KEL:94}.  Here, we
focus on the use of knowledge preconditions in accounting for the
initiation of information-seeking subdialogues.  We use the dialogue
in Figure~\ref{ac} as an example and assume the role of the Expert
(participant ``E'') in analyzing the discourse.\footnote{For
simplicity of exposition, we will take participant ``E'' to be female
and participant ``A'' to be male.}  The dialogue in Figure~\ref{ac}
was extracted from a larger discourse in which the Expert and
Apprentice (participant ``A'') are collaborating on removing the pump
of an air compressor.  We thus take the purpose of the larger
discourse to be

\begin{footnotesize}
\begin{tabbing}
{}~~~DSP$_1$=$Int.Th(e,FSP(\{a,e\},remove(pump(ac1),\{a\})))$\footnotemark\\
{}~~~~~~where $ac1$ represents the air compressor the agents\\
{}~~~~~~~~are working on.
\end{tabbing}
\end{footnotesize}
\footnotetext{We have omitted the time parameters for simplicity of
exposition.}

\subsection{Accounting for the Initiation of New Discourse Segments}
\label{recog}
To make sense of an {\em utterance}, an agent must provide an
explanation for it in the form of an answer to the question, ``Why did
the speaker say that to me?'' \cite{Sidner-Israel:81}.  An OCP must
provide a similar explanation for an ICP's initiation of a new
discourse segment.  This explanation takes the form of an answer to
the question ``Why does the ICP want to engage in a segment with
purpose DSP$_j$ at this point in our discourse?''; i.e., ``How is
DSP$_j$ related to what we were talking about before?''  Subsidiary
relationships between SharedPlans provide the basis for modeling the
OCP's reasoning.

One plan is subsidiary to another if the completion of the first plan
contributes to the completion of the second.  The most basic example
of this relationship occurs within the FSP definition itself.  As
indicated in Figure~\ref{sp-reqts}, a full plan for an act $\alpha$
includes full plans for each subact in $\alpha$'s recipe as components
(requirements~(2c) and (3b)).  The plans for the subacts thus
contribute to the plan for $\alpha$ and are therefore subsidiary to
it.

Subsidiary relationships may also arise in response to the other
requirements of the FSP definition.  For example, as discussed in
Section~\ref{know-prec}, the BCBA operator used to model
requirement~(2b) specifies that to be able to perform an act $\alpha$,
an agent must (1)~have a recipe for $\alpha$ ($has.recipe$), (2)~be
able to identify the parameters of $\alpha$ ($has.sat.descr$), and
(3)~have satisfied the constraints associated with performing
$\alpha$.  The first of these requirements provides an explanation for
the first subdialogue in Figure~\ref{ac}.

The purpose of this subdialogue is represented as\footnote{We describe
a method for recognizing DSPs of this form elsewhere \cite{KEL:94}.}
\begin{footnotesize}
\begin{tabbing}
{}~~~~~DSP$_2$=\=$Int.Th(a,FSP(\{a,e\},$\\
\>~~$Achieve(has.$\=$recipe(a,$\\[-0.5ex]
\>\>$remove(flywheel(ac1),\{a\}),R))))$
\end{tabbing}
\end{footnotesize}
and can be glossed as ``the Apprentice
intends that the agents collaborate on his obtaining a recipe for the
act of removing the flywheel of the air compressor.''  To account for
the Apprentice's initiation of this subdialogue, the Expert must
determine the relationship of DSP$_2$ to the purpose of the agents'
preceding discourse, namely DSP$_1$. In this case, the Expert can
reason that the Apprentice wants to engage in the subdialogue to
obtain a recipe for the act of removing the flywheel so that he will
be able to perform that act as part of the agents' SharedPlan to
remove the pump.  The plan in DSP$_2$ is thus subsidiary to that in
DSP$_1$ by virtue of a knowledge precondition requirement of the
latter plan.

Figure~\ref{kp1-graphic} illustrates this analysis.  Each box in the
figure corresponds to a discourse segment and contains the SharedPlan
used to model the segment's purpose.  The SharedPlans are labeled so
as to be co-indexed with the DSPs discussed above.  The arrows
indicate subsidiary relationships between SharedPlans, as explained by
the text that adjoins them.  When plan Pj is subsidiary to plan Pi,
DSP$_j$ is dominated by DSP$_i$.

\begin{small}
\begin{figure*}[bt]
\hspace*{1in}
\psfig{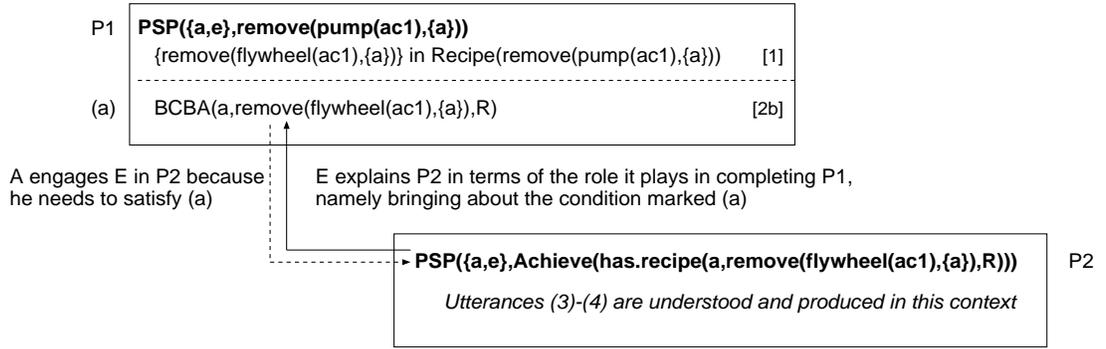}
\caption{Analysis of the First Subdialogue in
Figure~\protect\ref{ac}\label{kp1-graphic}}
\end{figure*}
\end{small}

The information represented within each SharedPlan in
Figure~\ref{kp1-graphic} is separated into two parts.  Those beliefs
and intentions that have been established at the time of the analysis
are shown above the dotted line, while those that remain to be
established, but that are used in determining subsidiary
relationships, are shown below the line.  The index in square brackets
to the right of each constituent indicates the FSP requirement from
which the constituent arose.

As indicated in Figure~\ref{kp2-graphic}, the initiation of the second
subdialogue in Figure~\ref{ac} is explained similarly.
This time, however, it is the need to identify parameters of acts
(requirement~(2) above) that leads to the initiation of the
subdialogue.  In addition, the parameter in question is a parameter of
an act in a subsidiary individual plan of the Apprentice's.

\begin{small}
\begin{figure*}[tb]
\hspace*{1in}
\psfig{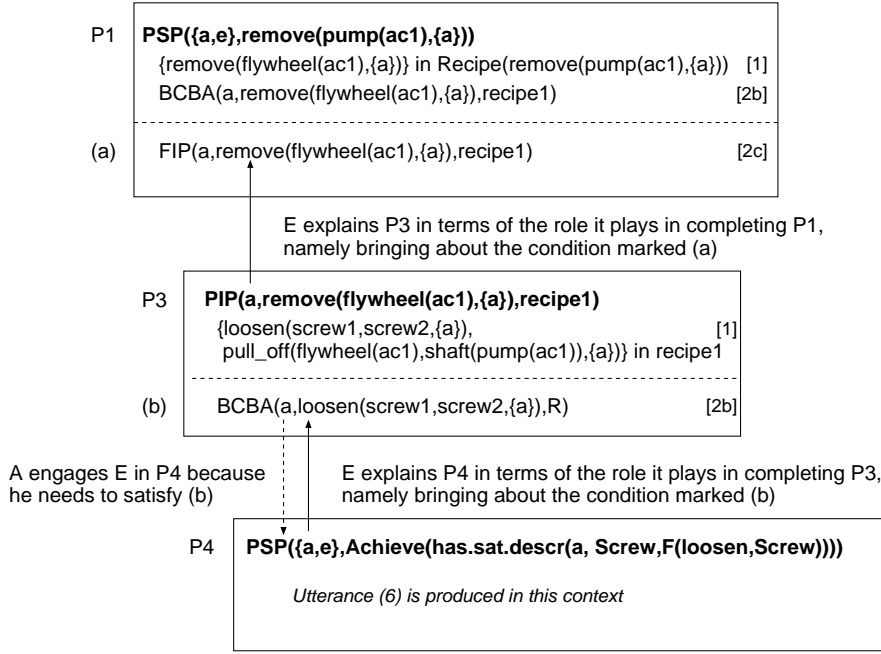}
\caption{Analysis of the Second Subdialogue in
Figure~\protect\ref{ac}\label{kp2-graphic}}
\end{figure*}
\end{small}

\subsection{Discussion}
In this paper, we have shown that information-seeking subdialogues may
be explained on the basis of knowledge precondition requirements.  Our
account of such subdialogues fits within a general framework for
discourse processing in which the purpose of a subdialogue is modeled
using a SharedPlan and is related to the purposes of other
subdialogues based on the requirements of the FSP definition.
Elsewhere~\cite{KEL:94}, we show that correction and subtask
subdialogues, among others, may also be accounted for in this manner.

In contrast, alternative plan-based accounts of dialogue understanding
introduce multiple types of plans to account for the utterances in a
discourse.  For example, Litman and Allen~\shortcite{Litman-Allen:87},
propose the use of two types of plans to model clarification and
correction subdialogues: {\em discourse plans\/} and {\em domain
plans}.  Domain plans represent knowledge about a task, while
discourse plans represent conversational relationships between
utterances and plans.  Litman and Allen provide operators for the
following discourse plans:
\begin{itemize}
\vspace*{-1ex}
\item INTRODUCE-PLAN: introduce a new plan for discussion
\vspace*{-1ex}
\item CONTINUE-PLAN: execute the next action in a plan
\vspace*{-1ex}
\item TRACK-PLAN: talk about the execution of an action
\vspace*{-1ex}
\item MODIFY-PLAN: introduce a new plan by modifying a previous one
\vspace*{-1ex}
\item CORRECT-PLAN: correct a plan
\vspace*{-1ex}
\item IDENTIFY-PARAMETER: identify a parameter of an action in a plan
\end{itemize}

Under our approach, the recognition of discourse plans is unnecessary.
The {\em fact} that a speaker is using an utterance to, for example,
introduce a plan, or track a plan, or identify a parameter, need not
be explicitly recognized for the purposes of utterance interpretation.
Furthermore, we would argue that such facts are not intended to be
recognized (cf. Grice~\shortcite{Grice:69}).  Rather, they simply fall
out of recognizing the relationship of an utterance to the current
discourse structure, i.e., the currently active SharedPlans.  For
example, INTRODUCE-PLAN corresponds to initiating a new discourse
segment, CONTINUE- or TRACK-PLAN to contributing to the current
segment, and IDENTIFY-PARAMETER to initiating a new segment to satisfy
a $has.sat.descr$ knowledge precondition requirement.  Although the
initiation of a new SharedPlan corresponds to the initiation of a new
discourse segment under our approach, it is the SharedPlan that must
be recognized and not a discourse plan that refers to that SharedPlan.

Lambert and Carberry~\shortcite{Lambert-Carberry:91} have extended
Litman and Allen's approach by introducing a third type of plan.  {\em
Problem-solving plans}, such as BUILD-PLAN and INSTANTIATE-VARS, are
used to model the process by which agents construct domain plans.
Under our approach, the need to explicitly recognize problem-solving
plans is also avoided.  The fact that an agent is building a plan or
instantiating a variable is a byproduct of understanding an utterance
by relating it to the current discourse structure.  BUILD-PLAN
corresponds to initiating a new discourse segment to satisfy a
$has.recipe$ knowledge precondition requirement, while
INSTANTIATE-VARS corresponds to initiating one to satisfy a
$has.sat.descr$ requirement.  Unlike Lambert and Carberry's approach,
however, and Litman and Allen's as well, our approach actually
recognizes this structure.  The other approaches are essentially
utterance-to-utterance based and thus do not recognize discourse
segments as separate units.

Ramshaw~\shortcite{Ramshaw:91} has added a different third type of
plan, {\em exploration plans}, to Litman and Allen's two types.
Exploration plans are intended to model the process by which agents
explore courses of actions.  Although we have not yet incorporated
such reasoning into our model, we hypothesize that the exploration of
plans can be modeled, without the introduction of a new plan type, by
reasoning about an agent's {\it potential intentions} and the process
by which they become full-fledged intentions
\cite{Grosz-Kraus:93,BIP:88}.

These alternative approaches share an important property that
distinguishes them from our approach; they take a {\em data-structure}
view of plans, rather than a {\em mental phenomenon} view
\cite{Pollack:90}.  Whereas data-structure plans are essentially
``recipes-for-action,'' mental phenomenon plans are a ``structured
collection of beliefs and intentions''
\cite[pg. 77]{Pollack:90}.\footnote{Although Lambert and
Carberry~\shortcite{Lambert-Carberry:91} adopt
Pollack's~\shortcite{Pollack:90} terminology in presenting their
theory, their ``plans'' are not mental state plans in Pollack's
sense.}  Data-structure plans thus describe {\em what} an agent is
doing with an utterance, but not {\em why} the agent is doing it.  For
example, although the constraints of Litman and Allen's
IDENTIFY-PARAMETER discourse plan force the plan to be related to
another plan that involves the parameter to be identified,
IDENTIFY-PARAMETER does not explain why this information is desired;
it does not capture that agents need to know parameters {\em to be
able to} perform acts involving them.\footnote{Although Litman and
Allen {\em describe} IDENTIFY-PARAMETER as providing ``a suitable
description for a term instantiating a parameter of an action such
that the hearer is then able to execute the action''
\cite[pg. 173]{Litman-Allen:87}, the IDENTIFY-PARAMETER operator
itself does not include a {\em formalization} of the last condition,
i.e., that the parameter description should enable the execution of
the action.}  It thus fails to model the essential knowledge
precondition nature of identifying a parameter.  Although it is
possible to impose a mental phenomenon interpretation on top of a
data-structure plan, doing so does not result in a mental phenomenon
plan \cite{Pollack:90}.  Saying that G1 {\em intends} to IDENTIFY a
PARAMETER fails to address why G1 intends to do so.

The need to explain an utterance is not unique to interpretation.
Moore and Paris~\shortcite{Moore-Paris:93} have shown that a similar
need exists in generation.  In particular, they have argued that
RST-based text plans must be augmented with intentional structure.
Otherwise, a system has no record of why it said what it did and is
thus unable to respond effectively if a hearer does not understand or
accept its utterances.

\section{Conclusion}
In this paper, we have presented an axiomatization of knowledge
preconditions for the SharedPlan model of collaborative activity
\cite{Grosz-Kraus:93}.  We have also shown how the requirements of
knowledge preconditions can be used to account for information-seeking
subdialogues in discourse.  Our account of this phenomenon fits within
a general framework for discourse processing in which SharedPlans and
relationships among them are used to model the intentional component
of Grosz and Sidner's~\shortcite{Grosz-Sidner:86} theory of discourse
structure.  Unlike the alternative approaches, our approach recognizes
and makes use of discourse structure.  In addition, it does not require
the introduction of new plan types.

\begin{small}

\end{small}

\end{document}